\documentclass[conference]{IEEEtran}

\IEEEoverridecommandlockouts
\usepackage{cite}
\usepackage{amsmath,amssymb,amsfonts}
\usepackage[ruled,linesnumbered,commentsnumbered]{algorithm2e}
\usepackage{multicol}
\usepackage{algorithmic}
\usepackage{graphicx}
\usepackage{tabularx}
\usepackage{textcomp}
\usepackage{xcolor}
\usepackage{float}
\usepackage[super]{nth}
\usepackage{subcaption}
\usepackage{booktabs}
\usepackage{multirow}
\usepackage{siunitx}
\usepackage{ragged2e}
\usepackage{threeparttable}
\usepackage{array}

\def\BibTeX{{\rm B\kern-.05em{\sc i\kern-.025em b}\kern-.08em
    T\kern-.1667em\lower.7ex\hbox{E}\kern-.125emX}}

\begin{document}
\renewcommand{\arraystretch}{1.1}
\title{A Transfer Learning Approach for Network Intrusion Detection}
\author{\IEEEauthorblockN{Peilun Wu\IEEEauthorrefmark{1},
Hui Guo\IEEEauthorrefmark{2} and Richard Buckland\IEEEauthorrefmark{3}}
\IEEEauthorblockA{School of Computer Science and Engineering,
University of New South Wales\\
Email: \IEEEauthorrefmark{1}z5100023@cse.unsw.edu.au,
\IEEEauthorrefmark{2}hui.g@unsw.edu.au,
\IEEEauthorrefmark{3}richardb@unsw.edu.au}}
\maketitle

\begin{abstract}
Convolution Neural Network (ConvNet) offers a high potential to generalize input data. It has been widely used in many application areas, such as visual imagery, where comprehensive learning datasets are available and a ConvNet model can be well trained and perform the required function effectively. 
ConvNet can also be applied to network intrusion detection.  However, the currently available datasets related to the network intrusion are often inadequate, which makes the ConvNet learning deficient, hence the trained model is not competent in detecting unknown intrusions. In this paper, we propose a ConvNet model using transfer learning for the network intrusion detection. The model consists of two concatenated ConvNets and is built on a two-stage learning process: learning a base dataset and transferring the learned knowledge to the learning of the target dataset. Our experiments on the NSL-KDD dataset show that the proposed model can improve the detection accuracy not only on the test dataset containing mostly known attacks (KDDTest+) but also on the test dataset featuring many novel attacks (KDDTest-21) -- about 2.68\% improvement on KDDTest+ and 22.02\% on KDDTest-21 can be achieved, as compared to the traditional ConvNet model.
\end{abstract}


\section{Introduction}\label{introduction}

ConvNet, as an effective deep learning solution for large scale data processing, has attracted increasingly attentions. More and more successful designs have been developed for varied applications, such as image recognition, text classification, and data extraction and regression. 

ConvNet can also be applied to the network intrusion detection (NID).
There are trillions of network transections and thousands of intrusions on the network each day. Each month, new attacks are created. With the increasing scale of the network and explosive number of users, the threats are growing.
Therefore, it is ultimately important to have a design that can effectively capture all attacks in time. 

So far, the ConvNet research for the network intrusion detection has mainly focused on learning algorithms. However, the quality of the datasets used in training ConvNet is also important, but has not drawn much attentions. According to
\cite{b1}, a significant number of researches on NID are still based on the original DARPA (1998-1999) and KDD 1999 datasets. These training and test datasets are highly redundant and contain many potential pitfalls \cite{b2,b3}, which makes the validation of the designs not very convincible. 

\begin{figure}[t]
\centerline{\includegraphics[width=\linewidth]{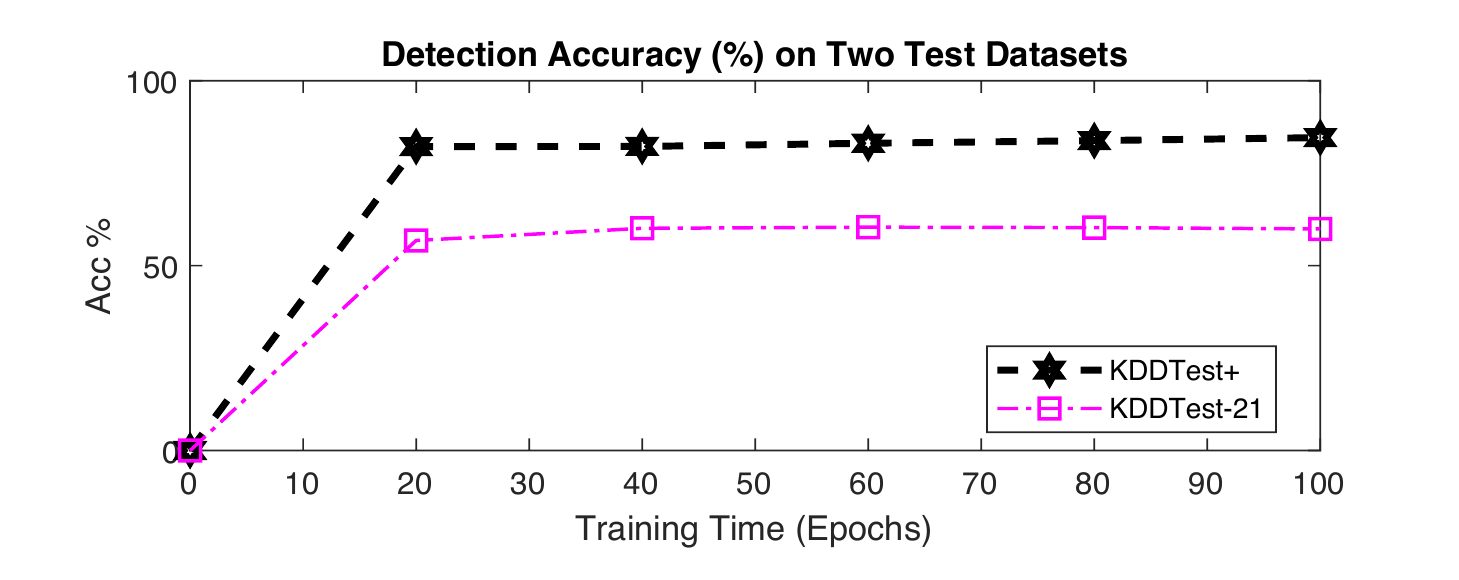}}
\caption{Large Performance Discrepancy of A Typical ConvNet}
\label{fig.motivation-example}
\end{figure}

Furthermore, even if a non-redundant dataset is used, a learning algorithm can demonstrate quite different performance on different test datasets. We have run an experiment to train a typical ConvNet model on the non-redundant NSL-KDD dataset. Fig. \ref{fig.motivation-example} shows the attack detection accuracies of the model that validate on two different test datasets: KDDTest+ and KDDTest-21. As can be seen from the figure, the model performs poorly on KDDTest-21.

We believe a major reason causing such an inferior performance is that the model is not sufficiently trained due to data deficiency of the training dataset.
A large portion of data in KDDTest-21 are attacks that are not covered in the training set of NSL-KDD \cite{b3}. 
Unfortunately, this situation of data deficiency in model training is quite common in practice, particularly for networks, where the attack types keep growing and evolving. {\bf How to build a ConvNet model on a limited learning dataset such that the model can perform intelligently and efficiently to detect network intrusions, both known and new?} This is the question we try to answer in this paper. 

Our solution to this problem is first gaining basic knowledge from some existing base dataset and learning the target dataset based on the acquired knowledge.

The contributions of our work are summarized as follows:

\begin{enumerate}

\item To our best knowledge, we are the first to address the issue of training data deficiency in using ConvNet for network intrusion detection. 

\item We introduce a knowledge-transfer based ConvNet model, TL-ConvNet. The model consists of two concatenated ConvNets. The first ConvNet holds the basic knowledge of network intrusions while the second one possesses the knowledge more specific to the target attack model.

\item We propose a novel training scheme to acquire the basic knowledge of network intrusions from a base dataset and 
train TL-ConvNet for the target dataset in two learning stages.

\item We build an experimental platform for the TL-ConvNet development and evaluation. Our experiment results demonstrate that the model is not only efficient in detecting known attacks but also much more effective in identifying new attacks that were not seen in the training dataset. 

\end{enumerate}

The rest of paper is organized as follows: Our TL-ConvNet model is presented in Section \ref{proposed framework} where the model structure and the training scheme are discussed. The experiments and results are given in Section \ref{performance evaluation} and 
a comparison to the related work on network intrusion detections can be found in Section \ref{related work}. The paper is concluded in Section \ref{conclusion}.\\

\section{TL-ConvNet Model}\label{proposed framework}

The intrusion attacks on the network can be performed in numerous and unforeseeable formats.  Hence, the signatures of attack may not be easily captured and effectively represented manually. However, a good data representation is critical to the final neural network (NN) created. 
Therefore, in our design, we use ConvNet (Convolution Network). The convolution operation in the ConvNet offers a powerful capability to extract high-level features of attack payloads and generate an effective representation of raw data automatically. 

Similar to other machine learning models, a ConvNet is trained on a {\bf training dataset} and tested on a {\bf test dataset}. 

We assume our ConvNet design targets a network intrusion model that contains known and novel attacks. We call the related training dataset {\bf target dataset}, and we regard the attacks covered by the training dataset {\bf known attacks} and those that appear only in the test dataset but not the training dataset {\bf novel attacks}. 

For our knowledge-transfer based ConvNet design, we further assume that there is an extra dataset that was created through a different data collection system and may have a different intrusion model. We use this dataset to pre-train our ConvNet model and the dataset is called {\bf base dataset}.

The design of our TL-ConvNet Model is elaborated in the next three subsections. 

\subsection{Model Structure}
 \begin{figure}[htbp]
    \centering
            \begin{subfigure}{\linewidth}
            \centering
            \includegraphics[width=.9\linewidth]{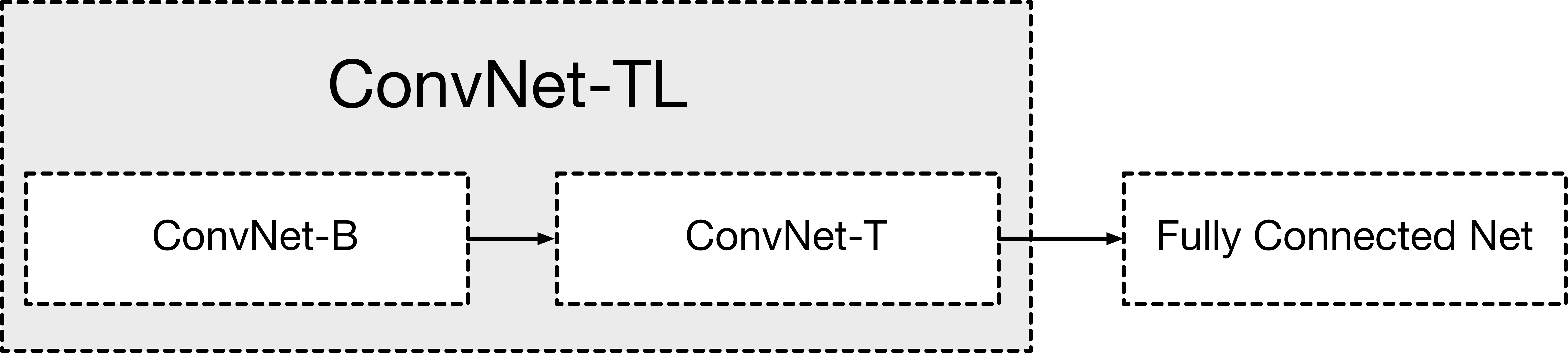}
            \caption{}
            \label{a}
            \end{subfigure}\\
            \begin{subfigure}{\linewidth}
            \centering
            \includegraphics[width=.9\linewidth]{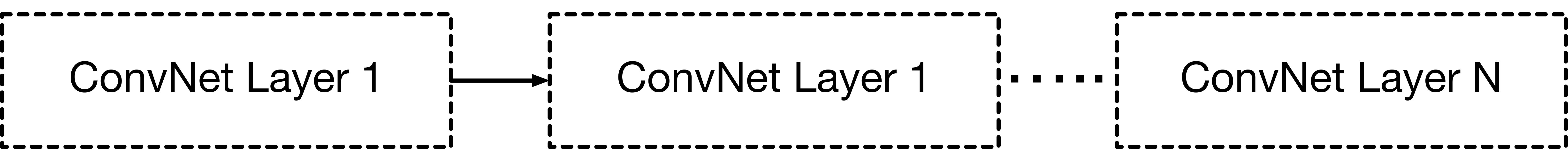}
            \caption{}
            \end{subfigure}\\
            \begin{subfigure}{\linewidth}
            \centering
            \includegraphics [width=.9\linewidth]{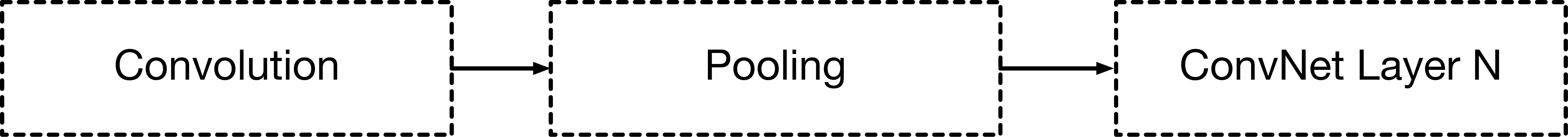}
            \caption{}
            \label{b}
            \end{subfigure}\\    
\caption{Structures: (a) TL-ConvNet Model; (b) ConvNet; (c) ConvNet Layer.}
\label{fig.model-structure}
\end{figure}
    
The general structure of the TL-ConvNet model is shown in Fig. \ref{fig.model-structure}(a). It contains two concatenated ConvNets (ConvNet-B and ConvNet-T) and a fully-connected net as an output layer. The combined two ConvNets is also named as ConvNet-TL.

Each ConvNet of ConvNet-TL  in turn contains multiple (typically two) ConvNet layers, as shown in Fig. \ref{fig.model-structure}(b), and each ConvNet layer is further constructed with three computation layers (as can often be observed in a typical ConvNet design): Convolution, Pooling, and Dropout (see Fig. \ref{fig.model-structure}(c)).

The convolution can be performed on 1D or multiple dimension data, depending on applications. For the network intrusion, the data items in the dataset often come from network packets and readily presented in vectors.
Therefore, it is straightforward to use 1D convolution in our design\footnote{Some existing designs such as\cite{b14} convert the raw data into images, which, we believe, is not necessary and may even introduce unexpected data loss.}.

Pooling in the ConvNet layer down-samples the output from the convolution. It can select the most active and useful features of the data to facilitate the next step learning \cite{b4}.

Dropout is randomly (on a fixed probability) to remove some connections from the NN layer to reduce possible overfitting that leads to degraded performance on the test dataset \cite{b5}. We therefore, include those computing layers for the base structure of the ConvNet. 

The output layer maps the ConvNet-TL results to the final output. Generally, the output layer contains a fully-connected net with an activation function to amplify the result.

We want to train the model for high attack detection accuracy on both known and novel attacks, which is explained below.

\subsection{Model Training}

\begin{figure}[t]
\begin{subfigure}{\linewidth}
\centerline{\includegraphics[width=.8\linewidth]{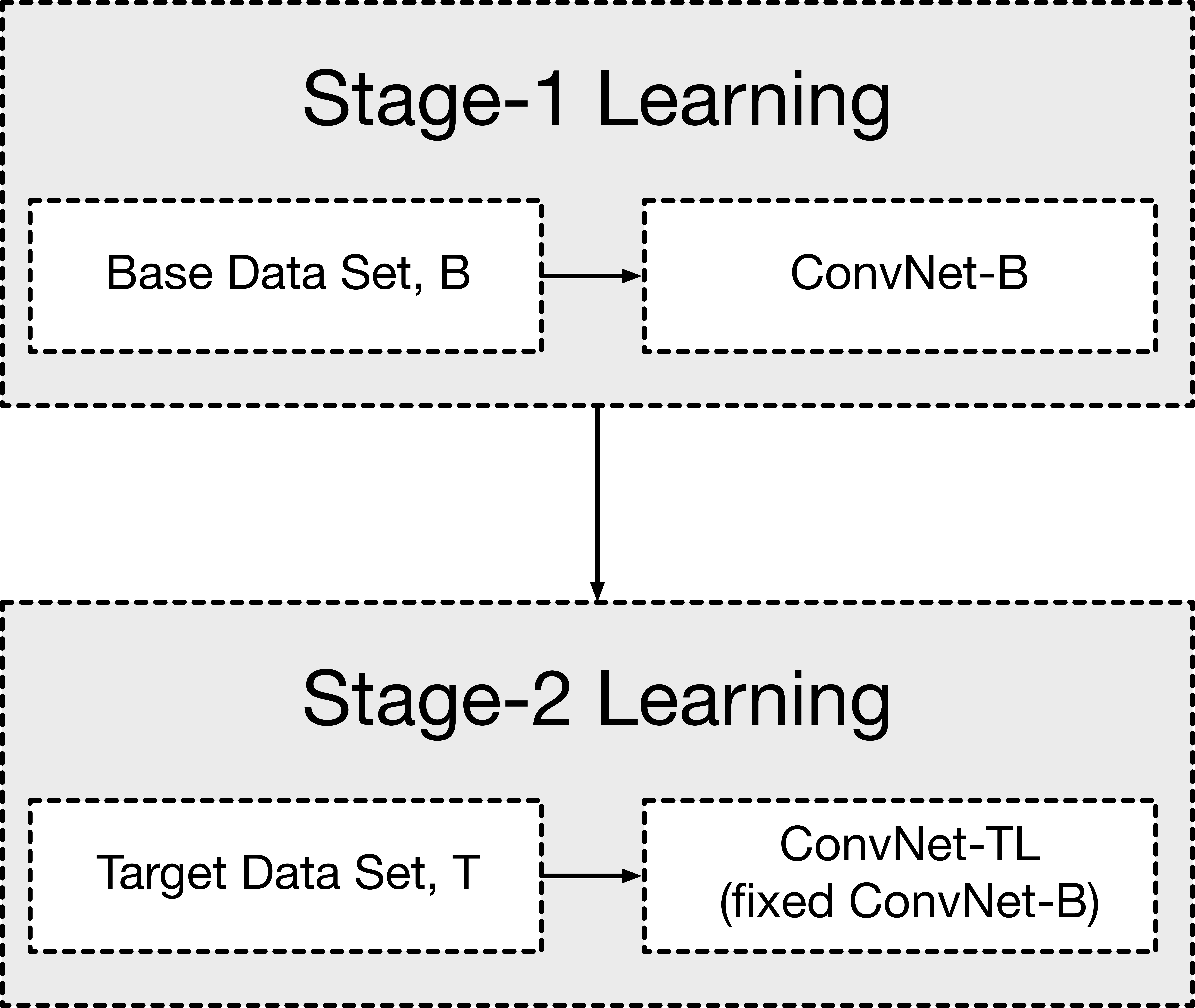}}
\caption{}
\end{subfigure}\\
\begin{subfigure}{\linewidth}
\centerline{\includegraphics[width=.8\linewidth]{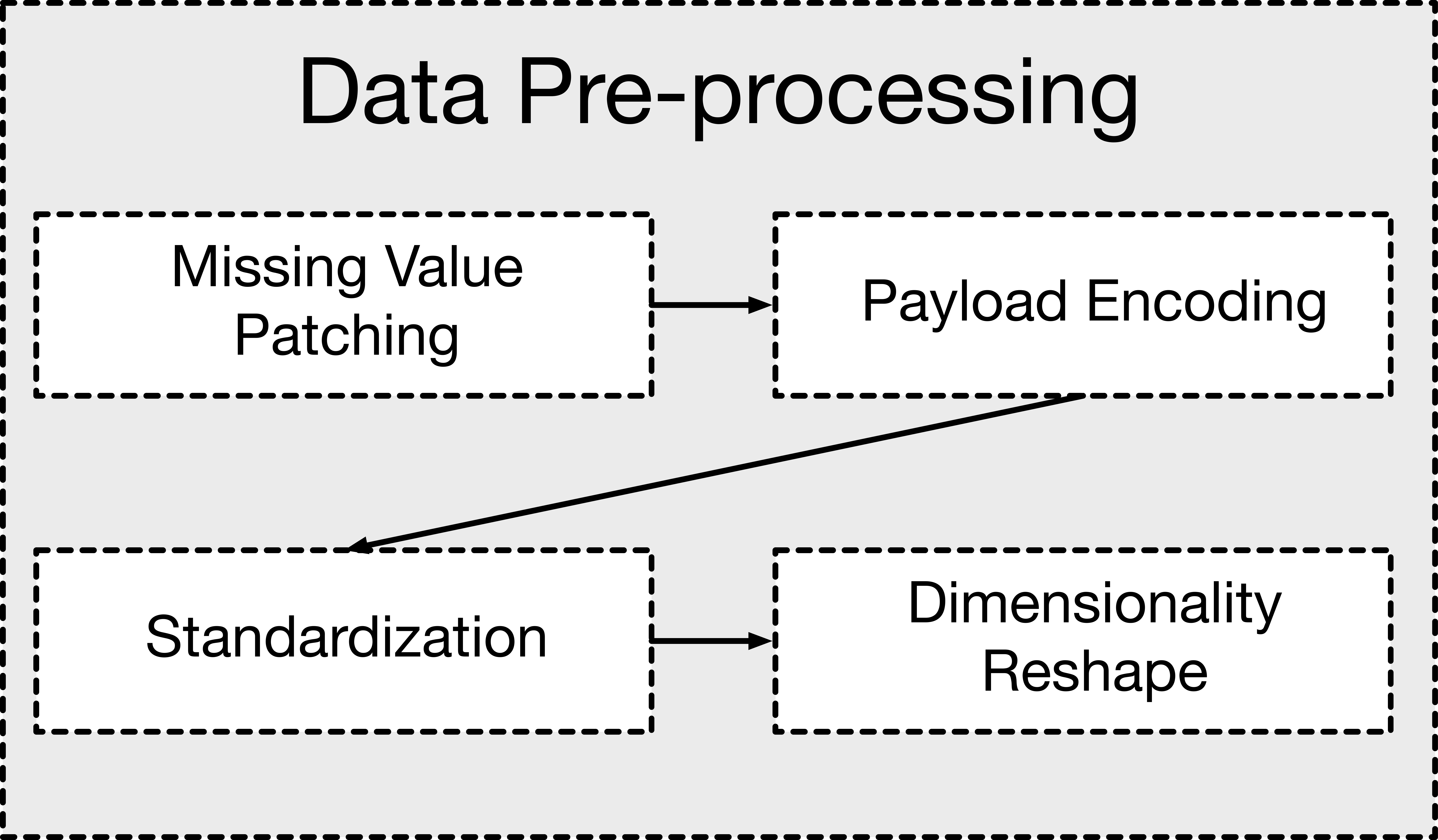}}
\caption{}
\end{subfigure}\\
\caption{Training: (a) TL-ConvNet Training Platform; (b) Data Preprocessing Steps.}
\label{fig.TL-model-training}
\end{figure}

\begin{figure}[t]
\centerline{\includegraphics[width=.9\linewidth]{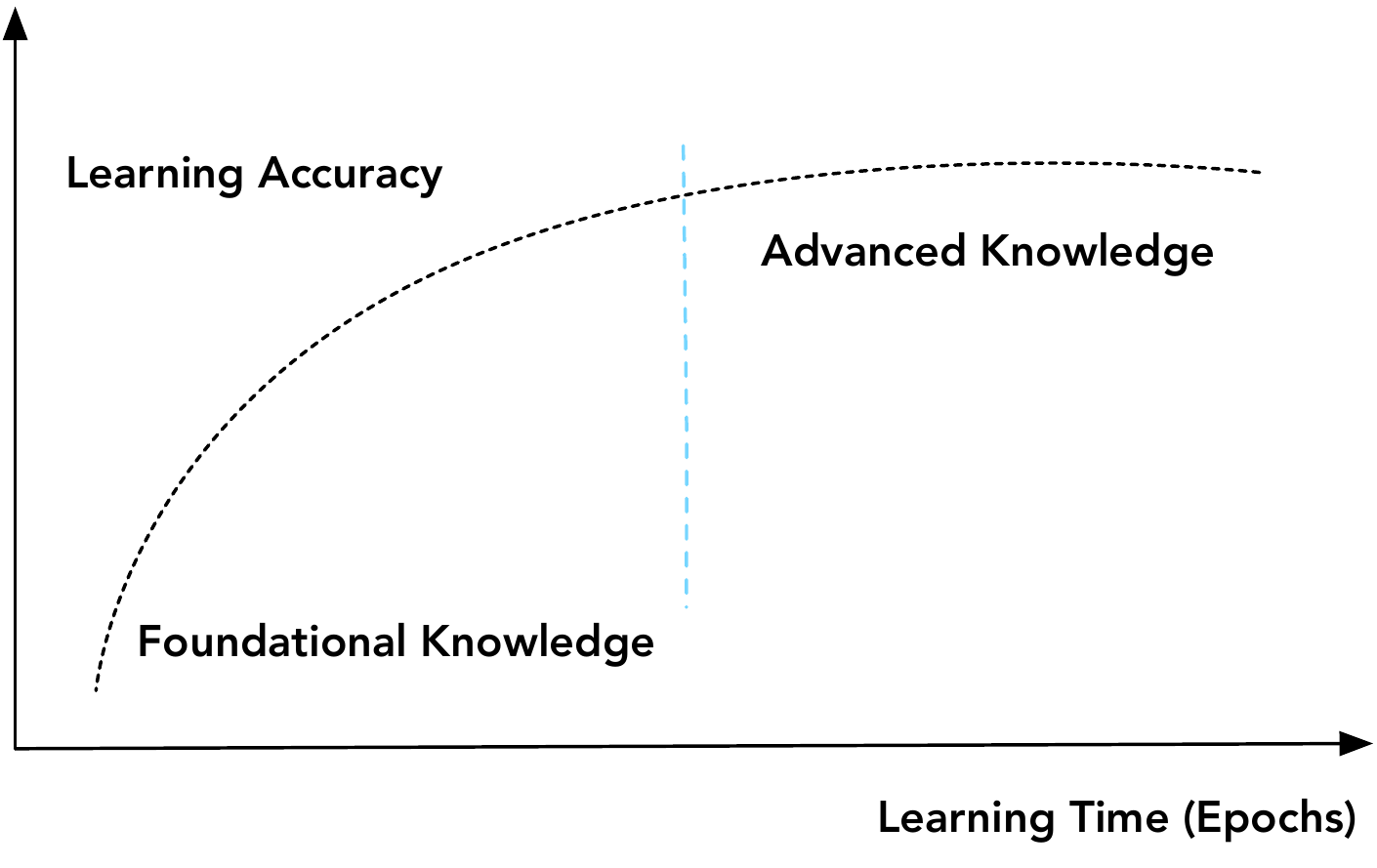}}
\caption{Knowledge versus Learning.}
\label{fig.Training-trend}
\end{figure}

The overview of training platform for TL-ConvNet Model is given in Fig. \ref{fig.TL-model-training}(a). 

It contains two main training stages: Stage-1 learning and Stage-2 learning. Like in all NN trainings, the learning is iterations of the net tuning for a target computing function.

Stage-1 learning trains ConvNet-B on the base dataset, $B$. The learning on the dataset aims to extract sufficient knowledge to help the learning of target dataset. 

After ConvNet-B has been trained, the second ConvNet, ConvNet-T, is added to form a whole ConvNet-TL model for Stage-2 learning.
In Stage-2 learning, ConvNet-B is fixed and only ConvNet-T is specifically trained on the target dataset, $T$. 

Once the training is converged with a stable detection accuracy, the model is finally built.

It must be pointed out that when training a ConvNet, an output layer (Fully Connected Layer) should be used to generate results for evaluation.  

Since no data collection system is perfect and the available raw data may not follow the ConvNet input format, the collected raw data should be preprocessed, as indicated in  Fig. \ref{fig.TL-model-training}(a) for each learning stage.

During the data preprocessing, data items with missing values are repaired with patches and all data are then converted into numerical values via encoding; The values are further scaled (standardized) to enhance the learning efficiency. Finally, the resulting data are reshaped into the format required for the ConvNet operation, as outlined in Fig. \ref{fig.TL-model-training}(b).

\subsection{Transferability}
One important issue with our TL-ConvNet model is how to ensure proper knowledge to be transferred from Stage-1 learning to Stage-2. 

For a ConvNet training, often  the learning accuracy increases with the training time, as demonstrated in Fig. \ref{fig.Training-trend}. 
We regard the knowledge gained during the early learning period is foundational. As the training progresses, the model gains more advanced knowledge about the training dataset, hence achieving higher learning accuracy. 

For our TL-ConvNet model, similar to some situations of using transfer learning for image classification \cite{b6}, the advanced knowledge to the base dataset may not be much relevant to the target dataset and may even adversely affect the learning accuracy of the overall model. 
Therefore, a small training time should be set to control its transferability.



Other parameters required for the TL-ConvNet training need to be further tuned for a good training result, which will be covered in the next section.

\section{Experiments and Results}\label{performance evaluation}

We built an evaluation platform on the HP EliteDesk 800 G2 SFF Desktop with Intel (R) Core (TM) i5-6500 CPU @ 3.20 GHz processor and 16.0 GB RAM. The ConvNet models are implemented based on the TensorFlow backend, and the frontend Keras and scikit-learn packages. 

\begin{table*}[htbp]
\newcommand{\baselinestratch}{1}
\centering
\caption{THREAT MODELS}
\begin{tabularx}{\linewidth}{lX}
\toprule
DATASET & ATTACK TYPES\\
\midrule
UNSW-NB15&ANALYSIS, BACKDOOR, DoS, EXPLOITS, FUZZERS, GENERIC, RECONNAISSANCE, SHELLCODE, WORMS.\\
NSL-KDD(TRAIN)&BACK, LAND, NEPTUNE, POD, SMURF, TEARDROP, SATAN, IPSWEEP, NMAP, PORTSWEEP, GUESS\_PASSWD, FTP\_WRITE, IMAP, PHF, MULTIHOP, WAREZMASTER, WAREZCLIENT, SPY, BUFFER\_OVERFLOW, LOADMODULE, ROOTKIT, PERL.\\
NSL-KDD(TEST)&BACK, LAND, NEPTUNE, POD, SMURF, TEARDROP, SATAN, IPSWEEP, NMAP, PORTSWEEP, GUESS\_PASSWD, FTP\_WRITE, IMAP, PHF, MULTIHOP, WAREZMASTER, WAREZCLIENT, SPY, BUFFER\_OVERFLOW, LOADMODULE, ROOTKIT, PERL, APACHE2, MAILBOMB, PROCESSTABLE, UDPSTORM, SNMPGETATTACK, SNMPGUESS, NAMED, WORM, SENDMAIL, SQLATTACK, HTTPTUNNEL, XTERM, PS, XLOCK, XSNOOP, MSCAN, SAINT.\\
\bottomrule
\end{tabularx}
\label{table1}
\end{table*}

\begin{table}[t]
\centering
\caption{STATISTIC OF DATASETS}
\begin{tabularx}{\linewidth}{lccccc}
\toprule
 \multirow{2}{*}{DATASET} &  \multirow{2}{*}{NORMAL} & \multirow{2}{*}{ ATTACK }& \% of&\% of\\
 &  & & ATTACK & NOVEL ATTACK\\
\midrule
UNSW-NB15& 93,000& 164,673 & 63.9& -\\
KDDTrain+& 67,343& 58,630 & 46.5& -\\
KDDTest+& 9,711& 12,833 & 56.9&17.3\\
KDDTest-21& 2,152& 9,698 & 81.8&32.8\\
\bottomrule
\end{tabularx}
\label{table2}
\end{table}
\subsection{Datasets and Threat Models}\label{AA}

Our TL-ConvNet model targets NSL-KDD dataset. As mentioned earlier, it has two test datasets: KDDTest+ and KDDTest-21.

For the base dataset, we chose the UNSW-NB15 dataset. 
UNSW-NB15 is a non-redundant dataset with many contemporary attacks. Compared with its preceding intrusion detection datasets, it has a balanced distribution between normal and attack data items \cite{b7,b8}. 
There is no separate test dataset for UNSW-NB15. We, therefore, partitioned the dataset for both training and testing purposes.

The threat models formed by the datasets (UNSW-NB15, NSL-KDD for training, and NSL-KDD for testing) are shown in Table~\ref{table1}. As can be seen from the table, NSL-KDD and UNSW-NB15 have different threat models and there are 17 types of novel attacks in the test dataset of NSL-KDD.  

We also investigated the attacks distributed in the datasets used in our experiment.
Table~\ref{table2} shows the number of data items that are normal (Column 2) and the number of attacks (Column 3) in each dataset. The percentage of attacks in each dataset is given in Column 4 and the last column shows the percentage only for novel attacks. As can be seen from the table, KDDTest-21 contains relatively more novel attacks -- about a third of data in KDDTest-21 are novel attacks.


For the input raw data, we combined the data fields of the UNSW-NB15 and NSL-KDD datasets. After preprocessing, each sample data in the two datasets is a vector of 113 values.


\subsection{Experimental Results}

Following the training procedure given in Fig. \ref{fig.TL-model-training}(a), we carried out the training on different configuration settings by exploring the space of parameters. The final configurations for the TL-ConvNet training are summarized in Table~\ref{table3}.

In our experiments, all ConvNets have two ConvNet layers. For ConvNet-B, the number of outputs (convolution filters) for layer 1 is 256 and  for layer 2 is 512. The Pooling in each layer uses the Max pooling function and the Dropout uses 0.5 as the retaining probability. The training is performed on a learning rate of 0.001, batch size of 400, and learning time (epoch) is 1 as shown in the second row of Table~\ref{table3}. The configuration for ConvNet-T is given in the last row. 

\begin{table}[t]
\centering
\caption{TRAINING CONFIGURATIONS}
\begin{tabularx}{\linewidth}{lp{0.58cm}p{0.58cm}p{0.58cm}p{0.58cm}>{\centering}p{0.58cm}>{\centering}p{0.58cm}p{0.58cm}}
\toprule
\multirow{2}{*}{ConvNet} & \multirow{2}{*}{layer1} & \multirow{2}{*}{layer2} & \multirow{2}{*}{Pooling} & \multirow{2}{*}{Dropout} &learning rate& batch size &learning time \\
\hline
ConvNet-B & 256 & 512& Max & 0.5 & 0.001 & 400&1\\
ConvNet-T& 8 & 16 & Max & 0.5 & 0.001 & 1100&100\\
\bottomrule
\end{tabularx}
\label{table3}
\end{table}

\begin{table}[t]
\centering
\caption{A COMPARISON OF KNOWLEDGE TRANSFER WITH NON-TRANSFER}
  \begin{tabularx}{\linewidth}{lSSSSSS}
    \toprule
    \multirow{2}{*}{Model} &
      \multicolumn{3}{c}{KDDTest+} &
      \multicolumn{3}{c}{KDDTest-21} \\
      & {DR\%} & {ACC\%} & {FPR\%} & {DR\%} & {ACC\%} & {FPR\%}\\
      \midrule
     ConvNet & 78.35 & 84.62 & 7.11 & 54.36& 59.92& 15.06 \\
     TL-ConvNet*&56.25&71.98&7.22&41.99&46.82&31.41\\
     TL-ConvNet& 93.86 & 87.30 & 21.38 & 99.82 & 81.94 & 98.65 \\
    \bottomrule
  \end{tabularx}
\label{table5}
\end{table}

Based on the training configurations in Table~\ref{table3}, we used 10-fold cross-validation on the UNSW-NB15 for training ConvNet-B . To see how effective of foundational knowledge transfer, we also investigated the model when ConvNet-B was fully trained with advanced knowledge of UNSW-NB15.



For evaluation we also trained other two models and validated on KDDTest+ and KDDTest-21 respetively. The detection rate (DR\%), accuracy (ACC\%), and  false prediction rate (FPR\%) for the design without knowledge transfer (simply named as ConvNet), with the advanced knowledge transfer (named as TL-ConvNet*) and with the foundational knowledge transfer (TL-ConvNet) are given in the Table \ref{table5}. The table shows the superiority of the foundational knowledge transfer design – almost 99.82\% attacks in the KDDTest-21 can be detected. Majority of them are novel attacks.
With our experiments, 
after transferring foundational knowledge, the TL-ConvNet could correctly detect 12,045 out of 12,833 on the KDDTest+ and 9,680 out of 9,698 attacks on the KDDTest-21. Compared with the traditional ConvNet, our method can improve the accuracy around 2.68\% on the KDDTest+ and 22.02\% on the KDDTest-21.

Although TL-ConvNet can significantly improve the overall accuracy, it must be pointed that, TL-ConvNet presents a high false alarm rate (21.38\%) on the KDDTest+ and (98.65\%) on KDDTest-21. 
The unusual phenomenon is mainly caused by the inherent drawback of NSL-KDD data set, which are the uneven distribution and high attack density (51.9\% and 81.8\%) in two test sets.

\section{A Comparison of Related Work}\label{related work}

There have been many machine-learning based designs proposed. For the network intrusion detection, the classification algorithms are mostly relevant. Some typical ones are J48, Naive Bayes, NBTree, Random Forest (RF), Random Tree (RT), Multi Layer Perception (MLP) and Support Vector Machine (SVM).
Their detection accuracies on the NSL-KDD dataset have been evaluated in \cite{b3}, as copied in Table~\ref{table6} (the first seven entries in the table).
From the table, we can see that among all the seven algorithms, NBTree has a highest performance (with an accuracy of 82.02\%).
 
One problem with these classical designs is their limited ability of data generalization and hence restricted performance \cite{b9}. To achieve a better performance, researchers have recently turned to deep learning.
One deep learning model is the recurrent neural network (RNN) \cite{b10} proposed by Y. Chuan-long. The model has a low false prediction rate.
Another neural network mostly studied is the ConvNet. \cite{b11,b12,b13,b14,b15,b16}.
In \cite{b15}, Lin et al. presented a character-level ConvNet (Char-IDS)  for intrusion detections, where network traffic data are converted into characters before to be processed by the convolution network.
ReNet and GoogLeNet discussed in \cite{b16}, also use data conversion, but the conversion in the two designs transforms the network traffic data into images in order to use the existing ConvNet trained on an image dataset.
Their model for the network intrusion detection is then built on the trained ConvNet and the knowledge learned from the image dataset is transferred to the final model for the intrusion detection.
 
Our TL-ConvNet is similar to ReNet, GoogLeNet in that we also use an extra dataset to pre-train the ConvNet.
However, TL-ConvNet does not involve any data conversion to a different application domain. We believe that such a conversion is not necessary in the presence of ConvNet since ConvNet can learn well from the raw data, and most importantly, the data conversion may lead to some data loss and eventually degrade the learning efficiency. Furthermore, our design offers a flexibility of transferring knowledge at different learning level and we demonstrate that the design with the foundational knowledge transferred is very effective. Compared with other designs listed in Table~\ref{table6}, our design has a highest detection accuracy on each of the two test datasets: KDDTest+ and KDDTest-21.


\begin{table}[t]
\centering
\begin{threeparttable}[htbp]
\centering
\caption{A COMPARISON OF RELATED WORK}
 \begin{tabularx}{\linewidth}{lp{0.7cm}p{0.7cm}p{0.7cm}p{0.7cm}p{0.7cm}p{0.7cm}}
    \toprule
    \multirow{2}{*}{Methods} &
      \multicolumn{3}{c}{KDDTest+} &
      \multicolumn{3}{c}{KDDTest-21} \\
      & {DR\%} & {ACC\%} & {FPR\%} & {DR\%} & {ACC\%} & {FPR\%}\\
      \midrule
    J48\cite{b3} & - & 81.05 & - & -& 63.97 & - \\
    NaiveBayes\cite{b3} & -& 76.56 & - & - & 55.77 & - \\
    NBTree\cite{b3} & - & 82.02 & -& - & 66.16& -\\
    RF\cite{b3}& -& 80.67 & -& - & 63.26& -\\
    DT\cite{b3} & - & 81.59 & -& - & 58.51& -\\   
    MLP\cite{b3} & - & 77.41 & -& - & 57.34& -\\
    SVM\cite{b3} & - & 69.52 & -& - & 68.55& -\\
    RNN\cite{b10} & 72.95 & 83.28 & 0.03& - & 68.55 & -\\
    Char-IDS\cite{b15} & 81.12 & 85.07 & 9.71 & -& 72.27 & -\\
    ReNet50\cite{b16} & 69.41 & 79.14 & 0.08& 99.63 & 81.57 & 99.81\\
    GoogLeNet\cite{b16} & 65.64 & 77.04 & 0.08 & 100 & 81.84 &100\\
    \textbf{TL-ConvNet} & \textbf{93.86} & \textbf{87.30} & \textbf{21.38} & \textbf{99.82} & \textbf{81.94} & \textbf{98.65} \\
    \bottomrule
  \end{tabularx}
\begin{tablenotes}
\item[*] we use '-' indicates that those evaluation metrics are not given in related literature.
\end{tablenotes}
\label{table6}
\end{threeparttable}
\end{table}

\section{CONCLUSION}\label{conclusion}

In this paper, we have addressed the problem with data deficiency in ConvNet training for the network intrusion detection. 
Existing machine learning based NID designs are not effective to detect the novel attacks. We proposed a knowledge transfer based ConvNet model that first learns the basic knowledge about network intrusion from a base dataset. 
Our experiment results on the NSL-KDD dataset shows that our design not only improves the detection accuracy on its KDDTest+ dataset but also greatly increases the detection accuracy on the KDDTest-21 dataset, which demonstrates that with the learned foundational knowledge, the model can efficiently learn the target dataset and gain high capability to identify novel attacks.



\end{document}